\pdfoutput=1

\documentclass[11pt]{article}

\usepackage{acl}

\usepackage{times}
\usepackage{latexsym}

\usepackage[T1]{fontenc}

\usepackage[utf8]{inputenc}

\usepackage{microtype}

\usepackage{inconsolata}

\usepackage{graphicx}
\usepackage{bm}
\usepackage{amsmath}
\usepackage{amsfonts}
\usepackage{bbm}

\usepackage{helvet}
\usepackage{courier}
\usepackage{multirow}
\usepackage{bm}
\usepackage{amsmath,amssymb}
\usepackage{mdwlist}

\usepackage{relsize}
\usepackage{amsfonts}
\usepackage{url}
\usepackage{graphicx}
\usepackage{subfigure}
\usepackage{caption}
\usepackage{hhline}
\usepackage{color}
\usepackage{colortbl}
\usepackage{booktabs}
\usepackage{tabularx}
\usepackage{makecell}
\usepackage[linesnumbered,boxed,ruled,commentsnumbered]{algorithm2e}
\usepackage{lipsum}
\title{\textsc{LongAgent}: Scaling Language Models to 128k Context through Multi-Agent Collaboration}

\author{Jun Zhao$^{1}$\footnotemark[1]\footnotemark[2]\;,\ \ Can Zu$^{1}$\footnotemark[1]\;,\ \ Hao Xu$^{1}$,\ \ Yi Lu$^{1}$,\ \ Wei He$^{1}$,\ \ Yiwen Ding$^{1}$,\\ \ \ \textbf{Tao Gui}$^{2}$\footnotemark[2]\;,\ \ \textbf{Qi Zhang}$^{1}$\footnotemark[2]\;,\ \ \textbf{Xuanjing Huang}$^{1}$\\
  $^1$School of Computer Science, Fudan University\\
  $^2$Institute of Modern Languages and Linguistics, Fudan University\\
  \texttt{\{zhaoj19,qz,tgui\}@fudan.edu.cn,czu22@m.fudan.edu.cn}}
\begin{document}
\maketitle
\renewcommand{\thefootnote}{\fnsymbol{footnote}}
\footnotetext[1]{Equal Contributions.}
\footnotetext[2]{Corresponding authors.}
\begin{abstract}
Large language models (LLMs) have demonstrated impressive performance in understanding language and executing complex reasoning tasks. However, LLMs with long context windows have been notorious for their expensive training costs and high inference latency.
Even the most advanced models such as GPT-4 and Claude2 often make mistakes when processing inputs of over $100k$ tokens, a phenomenon also known as \textit{lost in the middle}.
In this paper, we propose \textsc{LongAgent}, a method based on multi-agent collaboration, which scales LLMs (e.g., LLaMA) to a context of 128K and demonstrates potential superiority in long-text processing compared to GPT-4. 
In \textsc{LongAgent}, a leader is responsible for understanding user intent and directing team members to acquire information from documents.
Due to members' hallucinations, it is non-trivial for a leader to obtain accurate information from the responses of dozens to hundreds of members.
To address this, we develop an \textit{inter-member communication} mechanism to resolve response conflicts caused by hallucinations through information sharing.
Our experimental results indicate that \textsc{LongAgent} offers a promising alternative for long-text processing. The agent team instantiated with LLaMA-7B 
achieves significant improvements in tasks such as 128k-long text retrieval, multi-hop question answering, compared to GPT-4.

\end{abstract}

\section{Introduction}
Benefiting from increased model scales and massive pre-training corpus, large language models (LLMs) like GPT-4 \cite{openai2023gpt4} and LLaMA \cite{touvron2023llama} have continuously improved their capabilities. However, due to the inherent quadratic complexity of attention mechanisms, LLMs are often pre-trained with a limited context window to avoid unbearable computational costs. 
Once the input text length exceeds this limit, LLMs suffer from severe performance degradation \cite{xiao2023efficient,peng2023yarn}. This significantly constrains the effectiveness of LLMs in many practical applications, such as querying information from books, analyzing legal documents, or scientific papers.
    \begin{figure}[t]
        \includegraphics[width=\columnwidth]{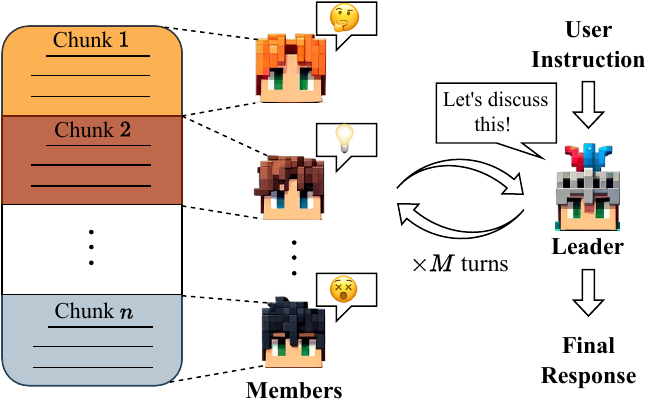}
        \caption{\textsc{LongAgent} collaboration scheme. The input long text (left) is segmented into several chunks and assigned to corresponding members. The Leader receives user instruction (right), breaks them down into the simplest sub-problems, convenes members for discussion, ultimately obtaining answers to all sub-problems, and reasons to make the final response.}
        \label{fig:intro}
    \end{figure}


Recently, many efforts have been dedicated to addressing the challenges of extending the context window of pre-trained LLMs. The first category of methods considers positional encoding as a crucial aspect to tackle this issue \cite{press2022train,chen2023extending,peng2023yarn,chen2023clex}. By extrapolation or interpolation, these methods enable positional encoding to adapt to positions ``unseen'' during the pre-training stage. 
However, this adaptation process may impair the performance on short sequences acquired during pre-training \cite{jin2024llm}. Additionally, as the window size increases, positional biases may decrease the effectiveness of attention mechanisms \cite{liu2023lost}, a phenomenon referred to as \textit{lost in the middle}.
The second category of methods employs intricate mechanisms such as designing recurrent structures \cite{zhou2023recurrentgpt,zhang2024soaring}, token selection \cite{mohtashami2023landmark,tworkowski2023focused}, or sliding windows \cite{xiao2023efficient,han2023lminfinite}, enabling limited context windows to handle longer input texts. While these methods offer advantages in computational efficiency, valuable long-term dependencies may gradually be overlooked with multiple recurrent iterations or window sliding.

In this work, we introduce a promising novel method, termed \textsc{LongAgent}, to address the challenge of handling long texts. \textsc{LongAgent} achieves processing of documents exceeding $100k$ through multi-agent collaboration. As illustrated in Figure \ref{fig:intro}, our agent team consists of a leader and multiple members. The leader is responsible for: 1) understanding user intent and organizing discussions among members; 2) supervising communication among members to resolve conflicting opinions; 3) gathering relevant information and reasoning answers. Each member is tasked with responding to the leader's instructions based on the content in the assigned text chunk. Due to members' hallucinations, it is non-trivial for a leader to obtain accurate information from the responses of dozens to hundreds of members. We address this issue through an inter-member communication mechanism. The leader detects members with conflicting opinions during discussions and allows them to interact directly to eliminate hallucinatory responses. In order to comprehensively evaluate LLMs' long-text capabilities, we further extend \textit{Needle in a Haystack}, a recently popular pressure test long-context LLMs. We change the simple fact retrieval to more challenging question-answering that may involve multiple documents.
The entities related to answers in the documents have been modified to prevent models from taking shortcuts by relying on internal knowledge. We named the new test \textit{Needle in a Haystack PLUS}\footnote{https://github.com/zuucan/NeedleInAHaystack-PLUS}. 

The main contributions of our
work are as follows: 1) we propose \textsc{LongAgent}, scaling LLMs with $4k$ context size to effectively handle long texts exceeding $100k$ tokens; 2) we construct a larger benchmark, \textit{Needle in the Haystack PLUS}, enabling more comprehensive evaluation on LLMs' long-text capabilities.
3) experimental results show that \textsc{LongAgent}, built upon LLaMA-7B, exhibits potential surpassing GPT-4 in long text processing. This improvement strongly suggests that multi-agent collaboration is a promising direction for improving long-text processing.

\section{\textsc{LongAgent} for Long Text Processing}
\subsection{Method Overview}
\begin{figure*}[t]
    \centering
        \includegraphics[width=\linewidth]{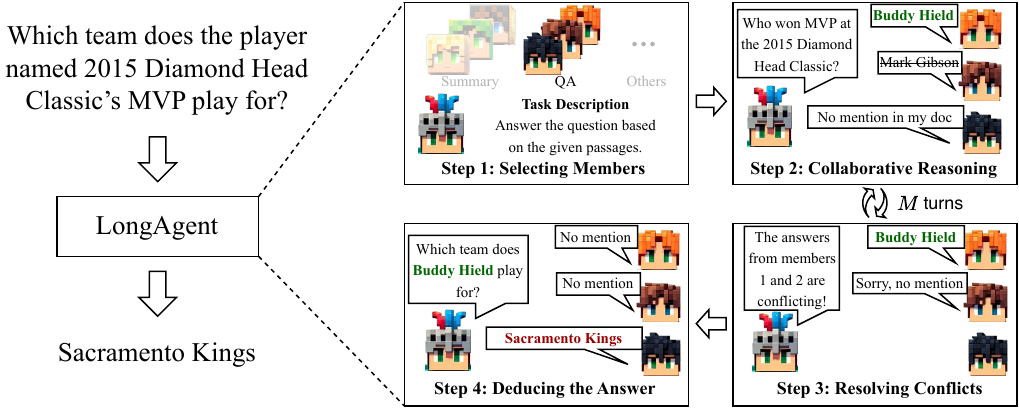}
        \caption{An Overview of the LongAgent. In step 1, the leader constructs a customized agent team based on the description of the task to be handled. In the second and third steps, the leader organizes the team to gather information from documents and resolve conflicts. This process may continue for multiple rounds until the leader deems enough information has been gathered to generate the final response, which is then exported in the step 4.}
        \label{fig:model}
    \end{figure*}

As illustrated in Figure \ref{fig:model}, we use long-text question answering as an example to elucidate the working mechanism of \textsc{LongAgent}. Given a long text $x$ and a user query $q$, \textsc{LongAgent} searches for relevant evidence from the long text $x$ and reasons for the final response $r$ through collaborative efforts of multiple agents. This process involves the following 4 steps: 

\noindent(1) \textbf{Selecting Members}. \textsc{LongAgent}, as a task-agnostic method for long-text processing, supports constructing customized agent teams based on the task to be processed. For example, given the task description $d=$``\textit{Answer the question based on the given passages}'', the leader selects the QA expert model to instantiate team members for more accurate and reliable responses. Then, we partition the long text $x$ into $n$ chunks $\{c_1, c_2, ..., c_n\}$ of predefined size and distribute them accordingly to $n$ members $\{m_1, m_2, ..., m_n\}$ for processing. 

\noindent(2) \textbf{Collaborative Reasoning}. For a complex user query $q$, the leader systematically breaks $q$ down into multiple sub-questions and organizes members to collaborate in searching for clues from their respective chunks. As shown in fig \ref{fig:model}, to answer $q =$ ``\textit{Which team does the player named 2015 Diamond Head Classic’s MVP play for?}'', the leader first identifies who won the MVP of the 2015 Diamond Head Classic, and then further identifies which team this player play for. For more complex queries, collaborative reasoning will involve multiple rounds. 

\noindent(3) \textbf{Resolving Conflict}. Due to model hallucinations, some members may respond with false answers not mentioned in the document. Obviously, these false answers conflict with the correct one. The leader needs to identify such conflicts and ascertain the correct answer explicitly mentioned in the document. 

\noindent(4) \textbf{Deducing the Answer}. When the leader believes that the current discussion history is sufficient to derive the answer to the user query, it ends the discussion and provide the final response. 

In the following sections, we will elaborate on the details of the aforementioned four steps.

\subsection{Selecting Experts to Instantiate Members}
The working mechanism of \textsc{LongAgent} involves coordinating members and integrating their processing results of respective chunks to form the final response. Different long-text tasks require members to possess diverse text processing capabilities. To address this challenge, we utilize expert models to construct task-specific agent teams, aiming to generate more accurate responses.

\noindent\textbf{Construction of expert models}: This step aims to build a candidate set of expert models $E=\{e_1, e_2,...,e_s\}$, where different experts excel in different tasks. For strong models (e.g., GPT-4, GPT-3.5), we employ a prompt-based approach to construct expert models. Specific prompt templates are used to elicit the internal capabilities of the model for the corresponding tasks. For example, ``\textit{You are an expert in answering questions, adept at searching for relevant information from given documents and providing answers}.'' The benefit of this approach is that all expert models can share the same parameters. For weak models (e.g., LLaMA-7b), we utilize a fine-tuning-based approach to inject task knowledge to make them experts in the corresponding tasks. The advantage of this method is that it can produce more accurate responses for specific tasks.

\noindent\textbf{Member selection}: Given a natural language description $d$ of a task to be processed, we prompt the leader to select a suitable expert $e_i\in E$ to play roles as team members. We assume that the task at hand requires only one particular expert to complete. For complex tasks that require collaboration among experts with different skill sets, we leave it as future work. The prompt template is illustrated in tab. \ref{prompt:single-doc_leader_first}. All members in the team share the parameters of the expert model $e_i$.

\subsection{Collaborative Reasoning}
To respond to user query $q$, the leader needs to coordinate members to process text and acquire relevant information. As the core of the team, the leader continuously executes the following decision-making process:
\begin{equation}
    a\sim \text{Leader}(a|S,q),
\end{equation}
where $q$ denotes the user query, $S = \{s_1, s_2,...,s_m\}$ represents the historical dialogue states, and $s_i$ represents a round of dialogue composed of an instruction from the leader and responses from all members. The leader sample an action $a\in \{$\texttt{NEW\_STATE}, \texttt{CONFLICT}, \texttt{ANSWER}$\}$  based on the dialogue history $S$ and the user query $q$. If $a = $ \texttt{NEW\_STATE}, it it indicates that the information contained in the preceding $i$ rounds of dialogue history is insufficient to respond to query $q$. Therefore, the leader initiates the next round of dialogue $s_{i+1}$, and generates new instructions to further gather information. Conversely, if $a = $ \texttt{ANSWER}, it signifies that the leader deems the currently collected information sufficient to derive the final answer, and the collaborative process terminates accordingly. \texttt{CONFLICT} is a special state other than the two aforementioned states, indicating that the leader perceives conflicting answers from the members in the current round of dialogue $s_i$. We elaborate on this situation in the next subsection.

\subsection{Resolving Conflicts}
\label{sec:conflicts}
Due to model hallucinations, members may respond with content not mentioned in their chunks. The dialogue in Step 2 of Figure \ref{fig:model} serves as an example, where two members respectively believe \textit{Buddy Hield} and \textit{Mark Gibson} to be the MVP of the 2015 Diamond Head Classic, despite the latter not being mentioned in the text chunk. We address this issue through \textit{inter-member communication}, inspired by the following empirical findings: 1) When there is answer to the leader's instruction in the chunk, the member often provides correct responses rather than generating hallucinations; 2) When there are no answers in the chunk, the model frequently fabricates an answer instead of responding with `no mention,' even after supervised fine-tuning. Using this feature, the leader first identifies the member IDs where answers conflict and then requests these members to share chunks pairwise and provide answers again:
\begin{gather}
    \text{hallucination}=m_i(c_i),\\
    \text{Truth}=m_j(c_j),\\
    \text{Truth}=m_j(c_j\oplus c_i)
\end{gather}
Here, $c_i$ and $c_j$ respectively represent two text chunks, where $c_j$ contains the correct answer while $c_i$ does not. $m_i$ and $m_j$ denote two members. Our experimental results demonstrate that sharing text chunks is a simple yet effective strategy. The majority of members experiencing hallucination tend to correct their original responses upon receiving the chunk containing the correct answers, resulting in accurate output. While we acknowledge some advanced mechanisms for mitigating hallucination issues, such as multi-agent debate \cite{du2023improving} and reflection \cite{shinn2023reflexion}, these are not the focus of this paper; we leave them as avenues for future research.

\section{Experimental Setup}
\subsection{Evaluation Protocol}
\noindent \textbf{Needle-in-a-Haystack PLUS}: The \textit{Needle-in-a-Haystack} \cite{NeedleInAHaystack} is currently one of the most popular testbed for evaluating the capability to handle long texts. In this setup, a fact or statement of interest (the \textit{needle}) is placed within a lengthy distracting document (the \textit{haystack}), and the model is tasked with retrieving this hidden key information. Performance is evaluated by varying the position of the needle within the distracting document and the length of the distracting document itself. To assess the long-text capabilities more comprehensively, we propose \textit{Needle-in-a-Haystack PLUS}, which shifts the focus from simple fact retrieval to more challenging single-document/multi-document question answering tasks. In \textit{Needle-in-a-Haystack PLUS}, the \textit{needle} represents the document(s) containing the answers, while the \textit{haystack} comprises distracting documents. The model must locate one or more relevant documents scattered within the haystack and reason the correct answers from them. For the purpose of evaluation, we deliberately select questions with definite answers as test data, such as questions where the answer is a specific entity or a simple yes/no response. To mitigate the risk of models relying on internal knowledge to answer, we replace entities directly related to the answer within the documents with fictional entities. In Appendix \ref{sec:needle_app}, we elaborate on the collecting process of the test data, including single-document QA and multi-hop QA involving multiple documents.

\noindent\textbf{Synthetic Tasks}: In addition to the \textit{Needle-in-a-Haystack PLUS} test, we also selected two widely used long sequence evaluation tasks \cite{mohtashami2023landmark,liu2023lost,zhang2023infinitebench}: long-text retrieval and numerical comparison. We choose them for the following reasons: (1) Similar to the needle-in-a-haystack task, these synthetic tasks all use ACC as the evaluation metric, facilitating evaluation. The difference in metrics can directly reflect the difference in the model's long sequence processing capabilities. (2) We can automatically synthesize training data for fine-tuning open-source models. The long-text retrieval task includes the following three subtasks: 1) PassKey Retrieval: Retrieving hidden keys in a noisy long context; 2) Number Retrieval: Locating repeated hidden numbers in a noisy long context. 3) KV Retrieval: Finding the corresponding value from a dictionary and a key.
Numerical comparison requires the model to find numbers that meet specific requirements from a numeric string of magnitude $100k$ tokens, such as the top $K$ numbers, median, etc., where $K$ can be $1$, $2$, or $3$.

\begin{figure*}[t]
    \centering
        \includegraphics[width=\linewidth]{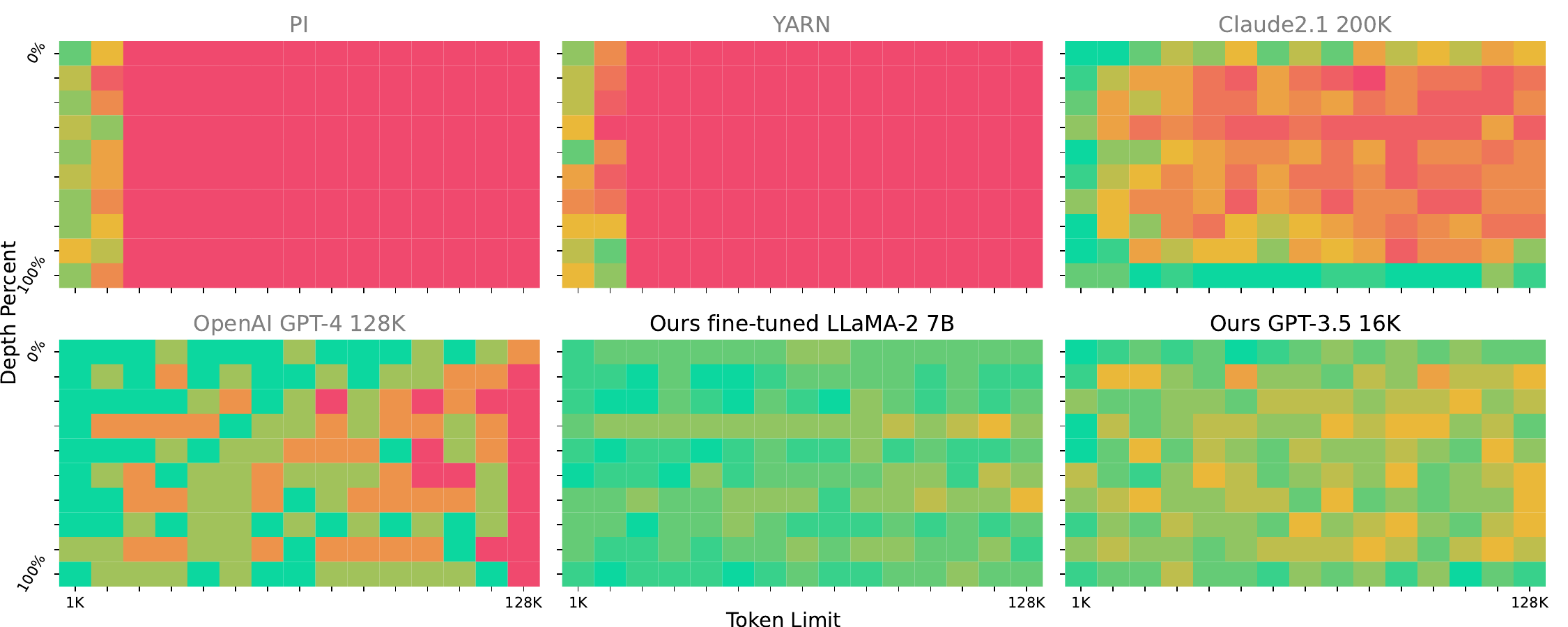}
        \caption{The Comparison of Results of \textit{Needle-in-a-Haystack PLUS} in Single-Document Question Answering Setting. Under the \textsc{LangAgent} scheme, our fine-tuned LLaMA2-7B model achieved an average accuracy improvement of $19.53\%$ compared to GPT-4 across the range from 1k to 128k (increasing from $62.00\%$ to $81.53\%$).}
        \label{fig:singleQARes}
    \end{figure*} 
    
\subsection{Compared Methods}
\noindent\textbf{PI} \cite{chen2023extending}. Extending the context window sizes of RoPE-based pretrained large language models by position interpolation.

\noindent\textbf{YARN} \cite{peng2023yarn}. YaRN is an improved method to efficiently extend the context window. This work directly modifies the PE to expand to a  theoretically infinite context length.

\noindent\textbf{Claude2.1} \cite{claude}. The Claude 2.1 released by Anthropic Corporation features a context window of 200K tokens and has significantly reductions in rates of model hallucination.

\noindent\textbf{GPT-4 Turbo} \cite{openai2023gpt4}. The GPT-4 Turbo model from OpenAI offers a context window of 128K and can process text exceeding 300 pages within a single prompt. 

\subsection{Implementation Details}
To build an agent team, we perform supervised fine-tuning on LLaMA2-7b-base. Within the agent team, the Leader is responsible for coordinating Members to accomplish various tasks. We utilize GPT-4 to generate $1,000$ interaction trajectories for each task to train the Leader, and manually verified the correctness of these interaction trajectories. Members are tasked with processing documents based on the Leader's instructions. To achieve this, we train QA experts, retrieval experts, and mathematical experts for instantiating members. Regardless of the number of members instantiated, they all share the parameters of a single expert model. Training data for QA experts are sourced from the SQuAD training set, consisting of $25,000$ samples. Among these, $10,000$ samples contain answers within the documents, while the remaining $15,000$ samples do not, requiring the model to abstain from answering. We extended document lengths to $2500$-$3000$ tokens through concatenation. Training data for retrieval experts and mathematical experts are synthesized automatically, with $10,000$ documents generated for each task, ranging in length from $1k$ to $3k$ tokens, and information to be retrieved evenly placed at random positions within the documents. It's important to note that all training data is non-overlapping with the final evaluation data. Please refer to Appendix \ref{sec:trajectory} for prompts and interaction trajectories for all tasks.
    
\section{Results and Discussion}
    \begin{figure*}[t]
    \centering
        \includegraphics[width=\linewidth]{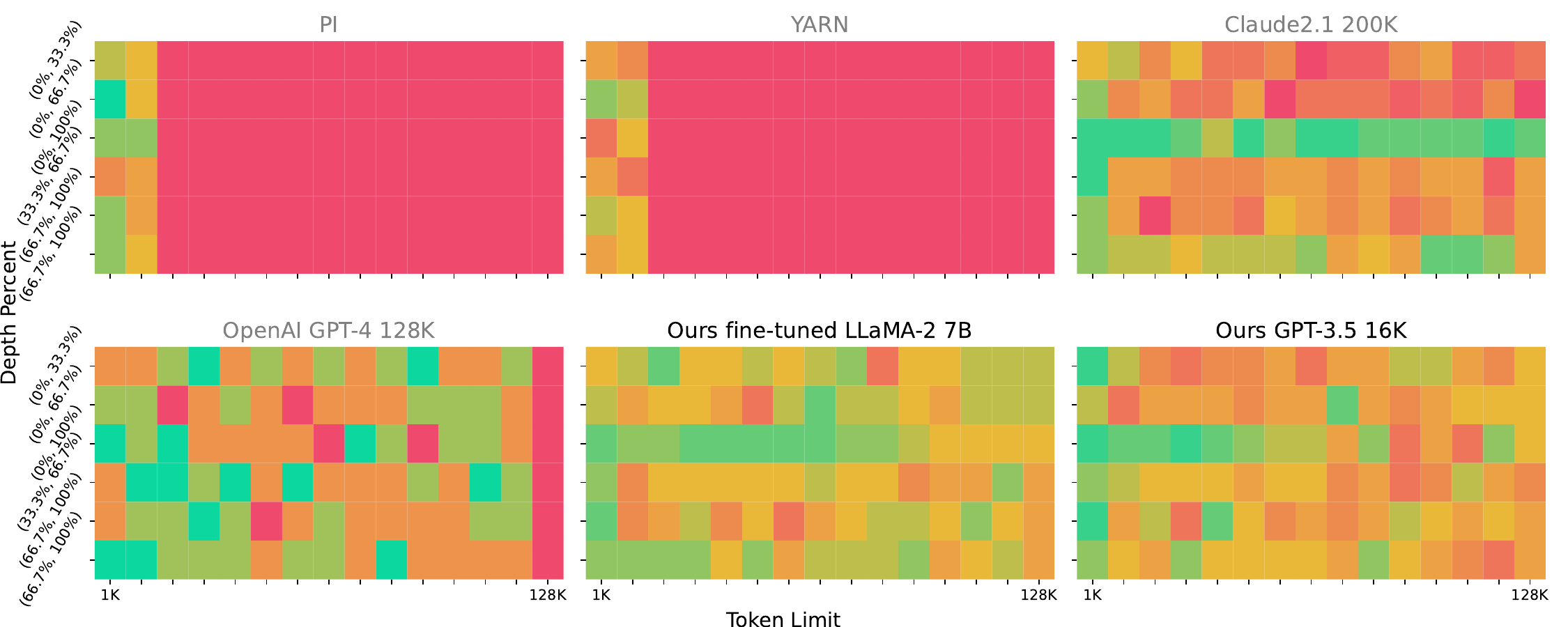}
        \caption{The Comparison of Results of \textit{Needle-in-a-Haystack PLUS} in Multi-Document Question Answering Setting. Under the LangAgent scheme, our fine-tuned LLaMA2-7B model achieved an average accuracy improvement of $4.96\%$ compared to GPT-4 across the range from $1k$ to $128k$ (increasing from $50.37\%$ to $55.33\%$).}
        \label{fig:multiQARes}
    \end{figure*} 

\subsection{Overall Performance}

To demonstrate the superiority of LongAgent in handling long texts, we compare it against powerful commercial models GPT-4 Turbo and Claude 2.1, as well as the state-of-the-art academic methods for long-text processing, PI and YARN. 

\noindent\textbf{Through multi-agent collaboration, fine-tuning LLaMA with only a 4k context window effectively handles contexts of up to 128k.}

The results for the \textit{Needle-in-a-Haystack PLUS} are shown in Figure \ref{fig:singleQARes} and \ref{fig:multiQARes}, respectively. \textsc{LongAgent}, constructed from fine-tuned LLaMA2-7B, significantly outperforms GPT-4 across document length ranging from 1k to 128k, with an average improvement of 19.53\% (from 62.00\% to 81.53\%) under the single-document setting, and an average improvement of 4.96\% (from 50.37\% to 55.33\%) under the multi-document setting. Considering that \textsc{LongAgent} is fine-tuned on downstream tasks, for fair comparison, we fine-tune PI and YARN on task data with lengths ranging from $1$ to $16k$ (training with longer contexts exceeds our hardware limitations). Experimental results demonstrate that when the length of the test document exceeds the maximum length trained on, PI and YARN fail to generate results properly. Even within the $0$-$16k$ range (corresponding to the first two columns of the grid), the average performance of \textsc{LongAgent} surpasses that of PI and YARN. The results on the four synthetic tasks are shown in Table \ref{tab:synthetic}. From the table, we can observe that \textsc{LongAgent} supported by fine-tuned LLaMA2-7B model outperforms all baseline models, achieving or approaching $100\%$ accuracy on the three retrieval-type tasks. This demonstrates the superiority of \textsc{LongAgent} in handling various long-text tasks.
    \begin{table}[t]
    \centering
    \resizebox{\linewidth}{!}{
    \begin{tabular}{l c cc c}
    \toprule
    \multirow{2}{*}{\textbf{Methods}} &  \multicolumn{3}{c}{Retrieval} & \multirow{2}{*}{\makecell[c]{Numerical\\ Comparison}}\\
    \cmidrule(r){2-4}
    & PassKey & Number & KV &\\
    \midrule
    GPT-4 & $\bm{1.000}$ & $\bm{1.000}$ & $0.890$ & $0.600$ \\
    Kimi-Chat & $0.981$ & $0.954$ & $0.536$ & $0.126$ \\ 
    Claude2.1 & $0.978$ & $0.981$ & $0.654$ & $0.323$\\
    YaRN & $0.927$ & $0.566$ & $-$ & $0.171$\\
    Ours-GPT3.5 & $\bm{1.000}$ & $\bm{1.000}$ & $0.638$ & $0.511$\\
    Ours-LLaMA-7B & $\bm{1.000}$ & $\bm{1.000}$ & $\bm{0.966}$ & $\bm{0.625}$\\
    \bottomrule
    \end{tabular}
    }
    \caption{The experimental results (accuracy) on four synthesis tasks.}
    \label{tab:synthetic}
\end{table}

    \begin{figure*}[t]
    \centering
        \includegraphics[width=\linewidth]{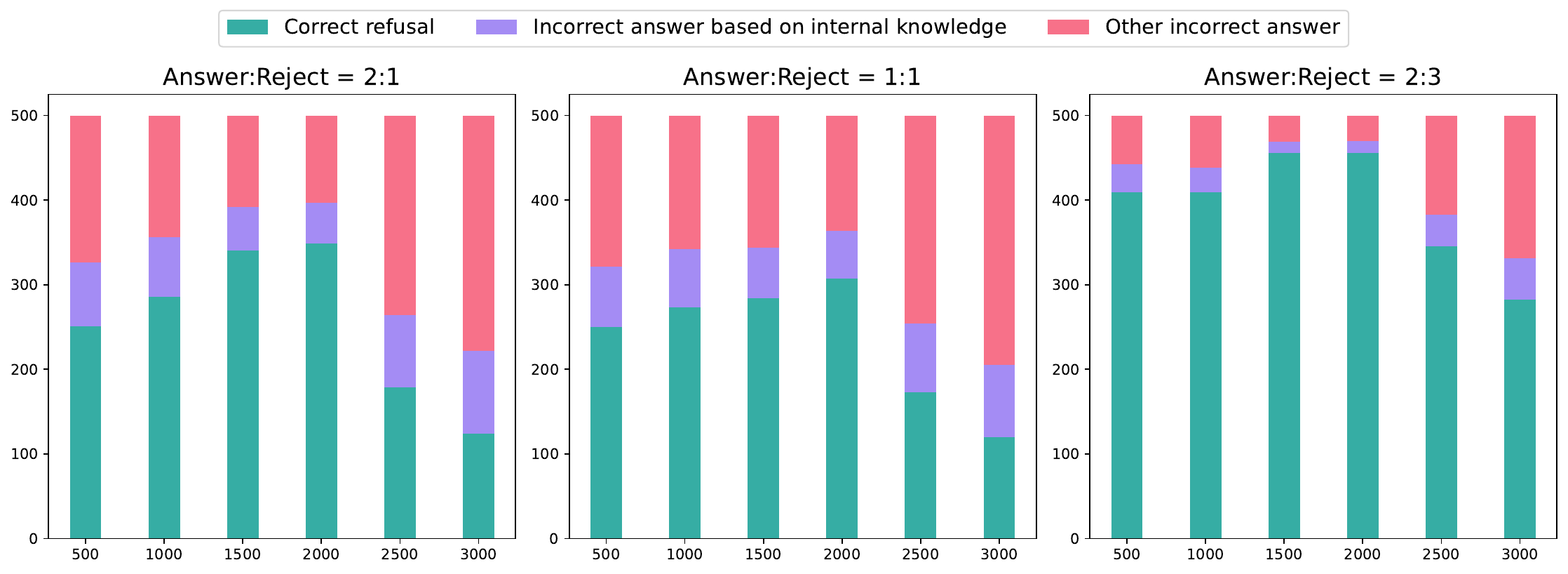}
        \caption{The influence of data recipe on model hallucinations. `Answer' and `Reject' represent two types of data. For the former, the documents contain answers to questions; whereas for the latter, they do not.}
        \label{fig:halluciantion}
    \end{figure*} 

\noindent\textbf{For \textsc{LongAgent} supported by more powerful models like GPT-3.5, fine-tuning is not necessary. }

Through prompting, GPT-3.5 can simultaneously act as a leader and members with specific skills. Despite having only a $16k$ context window, we found that the \textsc{LongAgent} supported by GPT-3.5 can effectively handle documents far exceeding $16k$ in length. Specifically, in the \textit{needle-in-a-haystack PLUS} task, \textsc{LongAgent} achieved improvements of $6.780\%$ and $1.5\%$ over GPT-4 in single-doc and multi-doc settings, respectively. For the four synthetic tasks in Table \ref{tab:synthetic}, \textsc{LongAgent} also achieved perfect scores in two retrieval tasks with $100k$ length documents. For KV retrieval and numerical comparison tasks, it also outperformed the majority of baselines.

\noindent\textbf{Although we only tested inputs ranging from 1k to 128k, \textsc{LongAgent} demonstrates potential in handling inputs exceeding 128k in length.}

In Figure \ref{fig:singleQARes} and \ref{fig:multiQARes}, we observed the \textit{lost in the middle} phenomenon with Claude 2.1. Specifically, as the input length increases, Claude2.1's average accuracy gradually decreases. However, in the first and last rows of the Claude subfigure in Figure \ref{fig:singleQARes}, and the third row of the Claude subfigure in Figure \ref{fig:multiQARes}, relatively high accuracy is consistently maintained. This suggests that Claude2.1 can effectively model the beginning and end of long texts but fails to utilize key information in the middle effectively. \textsc{LongAgent} avoids direct processing of long texts through chunking. Regardless of the input length, the chunk size remains constant, thus avoiding the 'lost in the middle' phenomenon. Although longer inputs may complicate agent interactions, experimental results show no significant performance decrease for \textsc{LongAgent}. Overall, \textsc{LongAgent} has the potential to handle inputs exceeding $128k$ in length.

\subsection{Hallucination Analysis}
We found that the errors of LongAgent are mainly due to a type of hallucination problem: when the chunk of a member does not contain information related to the instruction of the Leader, the member sometimes answers based on internal knowledge or fabricates a wrong answer as a response. In this subsection, we explore the impact of two key factors, the recipe of training data and chunk size, on model hallucination. As shown in Figure \ref{fig:halluciantion}, with the increase of 'Reject' type data in the training data, the proportion of the model correctly refusing to answer increased from $51.0\%$ to $78.6\%$. However, the increase of 'Reject' data also slightly impairs the model's ability to answer questions. As shown in Figure \ref{fig:hallucination2}, when the ratio of 'Answer:Reject' increases from 2:1 to 2:3, the accuracy of the model decreases from $83.3\%$ to $78.3\%$, and there are also a small number of cases where the document contains the answer but refuses to answer.

In addition to the data proportion, chunk size is also an important factor affecting model hallucination. As shown in Figure \ref{fig:halluciantion}, when the chunk size increases from $500$ to $2,000$, the hallucination problem is alleviated. This is mainly because the length of our training data is about $3,000$ tokens, and increasing the chunk size reduces the gap with the length of the training data. However, when the chunk size exceeds $2,000$, further increasing the chunk size significantly exacerbates model hallucination. We speculate that this is because when the sequence length is too long, the model's inadequacy in document modeling becomes more prominent. Therefore, we believe that while researching how to construct larger context windows, we should not neglect the modeling of text within a $4k$ window.
\subsection{Ablation Study}
    \begin{figure}[t]
        \includegraphics[width=\columnwidth]{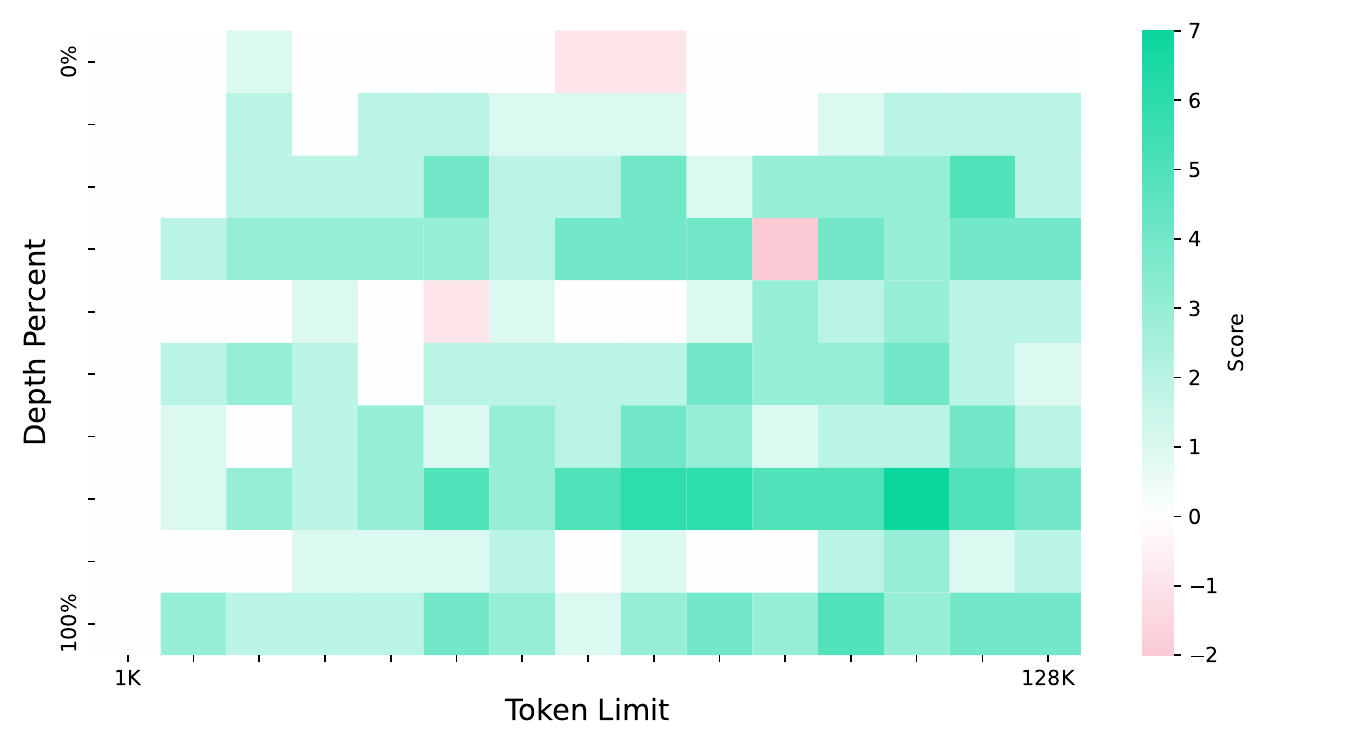}
        \caption{Improved accuracy through \textit{inter-member communication} mechanism.}
        \label{fig:ablation}
    \end{figure}
In Section \ref{sec:conflicts}, we address conflicts between members through \textit{inter-member communication}. To demonstrate the effectiveness of this mechanism, we calculate the difference in model accuracy before and after introducing this mechanism. As shown in Figure \ref{fig:ablation}, the \textit{inter-member communication} mechanism leads to an average accuracy improvement of $18.9\%$ across a range of input text lengths from $1k$ to $128k$. Furthermore, the number of members increases with the length of the text, and the number of members experiencing hallucinations also grows. In this context, the improvement in accuracy brought about by conflict resolution becomes even more evident.

\subsection{Efficiency Advantage}
    \begin{figure}[t]
        \includegraphics[width=\columnwidth]{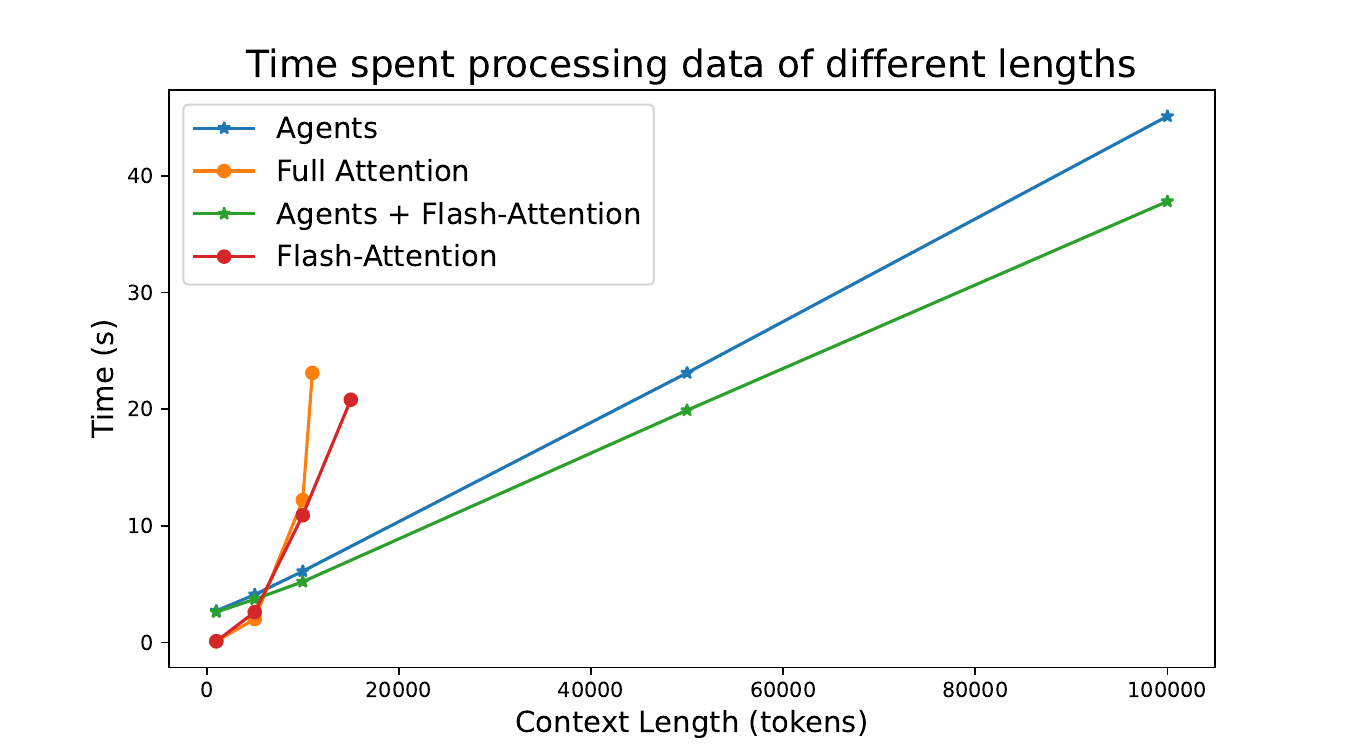}
        \caption{\textsc{LongAgent} scheme exhibits significantly superior time and memory efficiency compared to directly perform full attention on long texts.}
        \label{fig:latency}
    \end{figure}
Thanks to chunking of long texts, \textsc{LongAgent}'s time complexity for processing long texts is $\mathcal{O}(N)$. In this subsection, we empirically verify this point. As shown in Figure \ref{fig:latency}, the latency of \textsc{LongAgent} within the range of $1k$-$100k$ almost grows linearly with length. For Full Attention, which has quadratic complexity, the inference latency increases rapidly regardless of the use of techniques such as flash attention. The latency of Full Attention when processing 10k tokens has already exceeded that of \textsc{LongAgent} processing $50k$ tokens. Furthermore, without specific memory optimization techniques, a single A100 GPU with 80G memory can only support text inference up to 11k in length, and even with flash attention, this number can only be increased to $15k$. Under the same settings, \textsc{LongAgent} can process contexts of around $100k$ with less than 40G of memory.

\section{Related Works}
\subsection{Long-text Modeling}
Several methods have been proposed to extend the positional encoding (PE) for handling longer sequences. Initially, approaches like RoPE and PI \cite{chen2023extending} attempted to interpolate position indices within pre-trained limits, but neglected frequency variations. Recent advancements include "NTK-aware" \cite{bloc972023a} interpolation and "Dynamic NTK" \cite{bloc972023b} interpolation, which address high-frequency component losses. Additionally, "NTK-by-parts" \cite{bloc972023c} interpolation outperforms others when fine-tuned on longer-context data. Another popular approach for managing longer sequences involves constraining global causal attention to local attention. ReRoPE \cite{rerope2023} truncates context lengths during pretraining and LM-Infinite \cite{han2023lminfinite} restricts attention to a chevron-shaped window. \citet{mohtashami2023landmark} insert landmark tokens after text fragments, while \citet{zhang2024soaring} propose beacon tokens for summarizing fragments. In contrast, our method effectively circumvents the risk of losing valuable contextual information while utilizing only a small amount (hundreds of agent interaction tracks) for fine-tuning, thereby reducing training costs.
\subsection{LLM-Based Multi-Agent Systems}
In recent years, LLM-based multi-agent systems have garnered widespread attention in academia. Numerous efforts have been dedicated to leveraging cooperation among individuals to enhance the efficiency and accomplish more complex reasoning tasks \cite{du2023improving,wang2024unleashing,akata2023playing,hao2023chatllm}. To enable agents to effectively address a variety of dynamic tasks in the real world, researchers have also integrated external tools into the agents' decision-making processes \cite{cai2023large,gao2023assistgpt,paranjape2023art}, enabling them to perform accurate computations and retrieve the latest information from databases or search engines. 
In these approaches, the most relevant ones to ours are PEARL \cite{sun2023pearl} and MemWalker \cite{chen2023walking}. PEARL enhances the model's focus on relevant content within long texts by calling self-generated pseudo APIs. However, it can only handle long texts within the agent's context window and is ineffective for longer texts. Although MemWalker enables agents to process longer texts through a tree-based summarization approach, crucial information may be lost after multiple summarizations, causing the agent to get lost in irrelevant contexts. 
\section{Conclusions}
This paper proposes \textsc{LongAgent}, a novel long-text processing approach based on multi-agent collaboration. \textsc{LongAgent} scaling LLMs with $4k$ context size to effectively hadle long texts exceeding $100k$ tokens. The proposed \textit{inter-member communication} mechanism alleviates the member hallucination when they reading documents, thus facilitating effective management by the leader of dozens to hundreds of members. We have also developed \textit{Needle-in-a-Haystack Plus} to facilitate a comprehensive assessment of the LLM's capability with long texts. Our experimental results indicate that \textsc{LongAgent} offers a promising alternative for long-text processing.

\section*{Limitations}
\textsc{LongAgent} still has some drawbacks. Unlike general SFT data that only provides a prompt and a final response, \textsc{LongAgent}'s training data consists of interaction trajectories of multiple agents. Therefore, the construction cost of a single data point is higher, especially for tasks with more complex interaction trajectories. In addition, as the core of the agent squad, the Leader needs to make reasonable decompositions of the original complex problem and recruit members to solve it, which places higher demands on the Leader's reasoning and generalization abilities. For example, in the 'needle in a haystack' experiment, \textsc{LongAgent} improved by $19.53\%$ in a single-document setting compared to GPT-4, but this number dropped to $4.96\%$ when switching to a more complex multi-document setting. The main reason is that the reasoning ability of the LLaMA2-7B model is not sufficient to accurately decompose some complex problems.

\bibliography{custom}

\appendix
\newcommand{\graytexttt}[1]{\texttt{\textcolor{gray}{#1}}}
\section{Collecting Needle-in-a-Haystack PLUS Testing Data}
\label{sec:needle_app}
The test data consists of two parts: single-document QA and multi-document QA. Below, we will elaborate on the construction process of each.
\subsection{Single-document QA}
For single-document QA, the test data is constructed based on SQuAD \cite{rajpurkar-etal-2016-squad}. SQuAD is a large-scale machine reading comprehension dataset containing one hundred thousand questions, selected from over 500 Wikipedia documents. Each question in the dataset has its answer as a text segment extracted from the given document. We randomly select $100$ samples from the training set of SQuAD and replace the key entities in them. These $100$ samples' documents are used as \textit{needles}, while documents from other samples in the training set are randomly selected as \textit{haystacks}, which provide distractor information. The length of \textit{haystack} ranges from $1,000$ to $128,000$ tokens with equal intervals, totaling 15 different lengths. The depth of the needle ranges from $0\%$ to $100\%$ with equal intervals, totaling $10$ different depths. Depth indicates the position of the needle in the \textit{haystack}. A depth of $0\%$ means the \textit{needle} is at the beginning of the haystack, while $100\%$ indicates it's at the end. For each length and depth, we randomly select $10$ \textit{needles} to construct $10$ test samples.

\subsection{Multi-document QA}
For multi-document QA, questions require information from two or more documents to reason the final answer. We construct test samples based on HotpotQA \cite{yang2018hotpotqa}, a widely adopted multi-document QA dataset. We select $60$ questions from the validation set of HotpotQA that require information from two documents to answer. Each sample contains two \textit{needles}, and the haystack is still composed of distractor documents. The length of \textit{haystack} ranges from $1,000$ to $128,000$ tokens with equal intervals, totaling $15$ different lengths. The two \textit{needles} are randomly scattered at the depth of $0\%$, $33\%$, $66\%$, and $100\%$ of the haystack, resulting in $6$ combinations: $(0\%, 33\%)$, $(0\%, 66\%)$, $(0\%, 100\%)$, $(33\%, 66\%)$, $(33\%, 100\%)$, and $(66\%, 100\%)$. For specific lengths and needle positions, we randomly select $10$ \textit{needles} to construct $10$ test samples for evaluation.

\section{Trajectory for Each Task}
\label{sec:trajectory}

\subsection{Single-document Question Answering}
Dataset: Squad\\
\noindent\textbf{Leader first input}\\
First, the leader recruits members according to the task description as shown in Table \ref{prompt:single-doc_leader_first}.\\
\begin{table}[!ht]
\centering
\begin{tabular}{p{\linewidth}}
\toprule
You need to recruit a team of members to solve a task. Select the appropriate member based on the task description.\\
\textbf{\# Task Description:} \\
\{task\_description\}\\
\\
\textbf{\# Members List:} \\
QA member: Good at solving Question Answering problems.\\
KV member: Good at finding the corresponding value from a dictionary.\\
NS member: Good at locating repeated hidden numbers in a noisy long context.\\
PassKey member: Good at retrieving hidden keys in a noisy long context.\\
Math member: Good at finding special integers in a lengthy list.\\
\\
Your output must following the JSON format: \{\{"type": "member", "content": "your\_chosen\_member"\}\}\\
\bottomrule
\end{tabular}
\caption{Prompt template for Leader first input on the Single-document Question Answering Task. The content of \#Task Description is derived from user input.}
\label{prompt:single-doc_leader_first}
\end{table}\\
\noindent\textbf{Leader first output}\\
\textit{\{"type": "member", "content": "QA member"\}}\\
\noindent\textbf{Leader next input}\\
After recruiting members, the leader gives specific instruction as shown in Table \ref{prompt:single-doc_leader_next}.\\
\begin{table}[!ht]
\centering
\begin{tabular}{p{\linewidth}}
\toprule
You are the leader of a team of \{member\_nums\} members. Your team will need to collaborate to solve a task. The rule is:\\
1. Only you know the task description and task objective; the other members do not.\\
2. But they will receive different documents that may contain answers, and you need to send them an instruction to query their document.\\
3. Your instruction need to include your understanding of the task and what you need them to focus on. If necessary, your instructions can explicitly include the task objective.\\
4. Finally, you need to complete the task based on the query results they return.\\
\textbf{\# Task Description:} \\
Answer the question based on the given passages. The answer must be extracted from the given passages. \\
\\
\textbf{\# Task Objective:} \\
In which publication did Sander publish an article questioning racial preferences in law schools?\\
\\
\textbf{\# Generate Instruction for Members:} \\
Now, you need to generate an instruction for all team members. You can ask them to answer a certain question, or to extract information related to the task, based on their respective documents.\\
Your output must following the JSON format: \{\{"type": "instruction", "content": "your\_instruction\_content"\}\}\\
\bottomrule
\end{tabular}
\caption{Prompt template for Leader next input on the Single-document Question Answering Task. The content of \#Task Description and \#Task Objective is derived from user input.}
\label{prompt:single-doc_leader_next}
\end{table}\\
\noindent\textbf{Member first input}\\
The prompt for the Member first input is shown in Tabel \ref{prompt:single-doc_member_first}.
The content of \#Document is a part of the complete context, with variations for each member.
The content of \#Instruction originates from the first output of the leader.\\
\begin{table}[!ht]
\centering
\begin{tabular}{p{\linewidth}}
\toprule
\textbf{\# Document:} \\
\graytexttt{\{member\_document\}}\\
\\
\textbf{\# Instruction:} \\
Answer the question based on the given passages. The answer must be extracted from the given passages. Question: In which publication did Sander publish an article questioning racial preferences in law schools?\\
\\
You are an experienced writer; please summarize the content in the document related to the instructions in a <scratchpad> tag, then describe your response."
Your output must following the JSON format: \{\{"type": "response", "content": "your\_response\_content"\}\}\\
The "content" needs to be as concise as possible.\\
\bottomrule
\end{tabular}
\caption{Prompt template for Member first input on the Single-document Question Answering Task.}
\label{prompt:single-doc_member_first}
\end{table}
\noindent\textbf{Member first output}\\
The leader will exclude members who refuse to answer and group together members who provide the same answer.\\

\noindent\textit{Clusters: \{'the Wall Street Journal': [1, 7, 8, 13, 15, 17, 18, 22, 23, 24, 28, ,29 31, 32, 33, 34, 35, 37, 39, 42, 43], 'the Yale Law Journal': [5], 'The Wall Street Journal': [20]\}}\\
\\
\textbf{Member next input}\\
Prompt template for Member next input is the same as Member first input shown in Table \ref{prompt:single-doc_member_first}. The content of \#Document is the merge of the document from two members.\\
Select member 43 randomly from Cluster 1 and member 5 randomly from Cluster 2. First, merge the documents of member 43 and member 5 to obtain the new result.\\
\\
\textit{Member 43 and Member 5 merge: the Yale Law Journal}\\
\\
Cluster 1 is removed because its merged output matches the original output of member 5 but not member 43. Next, randomly select member 5 from Cluster 2 and member 20 from Cluster 3. Merge the documents of these two members to create the new result.\\
\\
\textit{Member 5 and Member 20 merge: Yale Law Journal}\\
\\
Remove Cluster 3 because the merged output is identical to the original output of member 5 and differs from the original output of member 20.\\
At this point, there is only one cluster left, thus all members have only one unique result. The leader will provide the final answer based on this result.\\
\noindent\textbf{Leader last input}
Leader needs to determine whether the final answer can be obtained. If not, further instructions will be provided. In this task, the final answer can be obtained. Details are shown in Table \ref{prompt:single-doc_leader_last}.\\
\begin{table}[!ht]
\centering
\begin{tabular}{p{\linewidth}}
\toprule
Here are the responses from all the members. Each member sees different segments of a document, and these segments do not intersect with each other. The correct answer may appear in any one or several members' responses.\\
Note that if a minority of members find information relevant to the question while the majority reply that the document does not contain information relevant to the question, you should pay attention to the replies from those members who found relevant information.\\
\textbf{\# Member Response:} \\
Member 1: Yale Law Journal \\
\\
\textbf{\# Task Description:} \\
Answer the question based on the given passages. The answer must be extracted from the given passages. \\
\\
\textbf{\# Task Objective:} \\
In which publication did Sander publish an article questioning racial preferences in law schools?\\
\\
\textbf{\# Determination:} \\
Based on the above information, you need to determine if you can solve the task objective. You have two choices:\\
1. If members' responses cannot solve the task objective, or if their responses contain conflicting answers, provide a new instruction for them to answer again.\\
2. Else, if the task objective can be solved, give your final answer as concisely as you can, using a single phrase if possible. Do not provide any explanation.\\
Your output must following the JSON format: \{\{"type": "answer", "content": "your\_answer\_content"\}\} or \{\{"type": "instruction", "content": "your\_instruction\_content"]\}\}\\
\bottomrule
\end{tabular}
\caption{Prompt template for Leader last input on the Single-document Question Answering Task.}
\label{prompt:single-doc_leader_last}
\end{table}\\
\noindent\textbf{Leader last output}\\
\noindent\textit{{"type": "answer", "content": "Yale Law Journal"}}\\

\subsection{Multi-document Question Answering}
Dataset: Hotpotqa\\
\noindent\textbf{Leader first input}\\
First, the leader recruits members according to the task description as shown in Table \ref{prompt:single-doc_leader_first}.\\
\noindent\textbf{Leader first output}\\
\textit{\{"type": "member", "content": "QA member"\}}\\
\textbf{Leader second input}\\
For multi-document Question Answering, the questions require answers that integrate the content of multiple documents. The details are shown in Table \ref{prompt:multi-doc_leader_second}.\\
\begin{table*}[!ht]
\centering
\begin{tabular}{p{\linewidth}}
\toprule
You are the leader of a team of \{member\_nums\} members. Your team will need to collaborate to solve a task. The rule is:\\
1. Only you know the task description and task objective; the other members do not.\\
2. But they will receive different documents that may contain answers, and you need to send them an instruction to query their document.\\
3. Your instruction need to include your understanding of the task and what you need them to focus on. If necessary, your instructions can explicitly include the task objective.\\
4. Finally, you need to complete the task based on the query results they return.\\
\textbf{\# Task Description:} \\
Answer the question based on the given passages. Only give me the answer and do not output any other words. \\
\\
\textbf{\# Task Objective:} \\
Did Leshan or Quanzhou  have a population of 658,798 in 2002? \\
\\
\textbf{\# Generate Instruction for Members:} \\
Now, you need to generate an instruction for all team members. You can ask them to answer a certain question, or to extract information related to the task, based on their respective documents.\\
Your output must following the JSON format: \{\{"type": "instruction", "content": "your\_instruction\_content"\}\}\\
\bottomrule
\end{tabular}
\caption{Prompt template for Leader second input on the Multi-document Question Answering Task. The content of \#Task Description and \#Task Objective is derived from user input.}
\label{prompt:multi-doc_leader_second}
\end{table*}
\noindent\textbf{Leader second output}\\
The leader needs to break down the original question, so the leader's second output is the first sub-question.\\
\textit{{"type": "instruction", "content": "What was the population of Leshan in 2002?"}}\\
\noindent\textbf{Member first input}\\
The prompt for the Member first input is shown in Tabel \ref{prompt:multi-doc_member_first}.
The content of \#Document is a part of the complete context, with variations for each member.
The content of \#Instruction originates from the first output of the leader.\\

\begin{table}[!ht]
\centering
\begin{tabular}{p{\linewidth}}
\toprule
\textbf{\# Document:} \\
\graytexttt{\{member\_document\}}\\
\\
\textbf{\# Instruction:} \\
What was the population of Leshan in 2002?\\
\\
You are an experienced writer; please summarize the content in the document related to the instructions in a <scratchpad> tag, then describe your response."
Your output must following the JSON format: \{\{"type": "response", "content": "your\_response\_content"\}\}\\
The "content" needs to be as concise as possible.\\
\bottomrule
\end{tabular}
\caption{Prompt template for Member first input on the Multi-document Question Answering Task.}
\label{prompt:multi-doc_member_first}
\end{table}
\noindent\textbf{Member first output}\\
\textit{Member 1: 121,277\\
Member 2: The document does not contain the answer.\\
Member 3: The document does not contain the answer.\\
\ldots\ldots\\
Member 30: The document does not contain the answer.\\
Member 31: The document does not contain the answer.\\}
\noindent\textbf{Leader third input}\\
The prompt for the Leader third input is shown in Tabel \ref{prompt:multi-doc_leader_third}.
At this moment, Leader still retains the dialogue history from the first round.
\begin{table*}[!ht]
\centering
\begin{tabular}{p{\linewidth}}
\toprule
Here are the responses from all the members. Each member sees different segments of a document, and these segments do not intersect with each other. The correct answer may appear in any one or several members' responses.\\
Note that if a minority of members find information relevant to the question while the majority reply that the document does not contain information relevant to the question, you should pay attention to the replies from those members who found relevant information.\\
\textbf{\# Member Response:} \\
Member 1: 121,277 \\
\\
\textbf{\# Task Description:} \\
Answer the question based on the given passages. Only give me the answer and do not output any other words. \\
\\
\textbf{\# Task Objective:} \\
Did Leshan or Quanzhou  have a population of 658,798 in 2002? \\
\\
\textbf{\# Determination:} \\
Based on the above information, you need to determine if you can solve the task objective. You have two choices:\\
1. If members' responses cannot solve the task objective, or if their responses contain conflicting answers, provide a new instruction for them to answer again.\\
2. Else, if the task objective can be solved, give your final answer as concisely as you can, using a single phrase if possible. Do not provide any explanation.\\
Your output must following the JSON format: \{\{"type": "answer", "content": "your\_answer\_content"\}\} or \{\{"type": "instruction", "content": "your\_instruction\_content"]\}\}\\
\bottomrule
\end{tabular}
\caption{Prompt template for Leader third input on the Multi-document Question Answering Task.}
\label{prompt:multi-doc_leader_third}
\end{table*}\\
\noindent\textbf{Leader second output}\\
\textit{\{"type": "instruction", "content": "What was the population of Quanzhou in 2002?"\}}\\
\noindent\textbf{Member second input}\\
The prompt for the Member second input is shown in Tabel \ref{prompt:multi-doc_member_second}.
The content of \#Document  is a part of the complete context, with variations for each member.
The content of \#Instruction originates from the second output of the leader.\\
\begin{table}[!ht]
\centering
\begin{tabular}{p{\linewidth}}
\toprule
\textbf{\# Document:} \\
\graytexttt{\{member\_document\}}\\
\\
\textbf{\# Instruction:} \\
What was the population of Leshan in 2002?\\
\\
You are an experienced writer; please summarize the content in the document related to the instructions in a <scratchpad> tag, then describe your response."
Your output must following the JSON format: \{\{"type": "response", "content": "your\_response\_content"\}\}\\
The "content" needs to be as concise as possible.\\
\bottomrule
\end{tabular}
\caption{Prompt template for Member second input on the Multi-document Question Answering Task.}
\label{prompt:multi-doc_member_second}
\end{table}
\noindent\textbf{Member second output}\\
Exclude the members who refuse to respond, and cluster the members who provide the same answers.\\

\textit{Clusters: \{'1.5 million': [7], '658,798': [11]\}}\\

\noindent\textbf{Member third input}\\
The prompt for the Member third input is shown in Tabel \ref{prompt:multi-doc_member_third}.
The content of \#Document is the merge of the document from Member 7 and Member 11. \\
\begin{table}[!ht]
\centering
\begin{tabular}{p{\linewidth}}
\toprule
\textbf{\# Document:} \\
\graytexttt{\{member\_document\}}\\
\\
\textbf{\# Instruction:} \\
What was the population of Quanzhou in 2002?\\
\\
You are an experienced writer; please summarize the content in the document related to the instructions in a <scratchpad> tag, then describe your response."
Your output must following the JSON format: \{\{"type": "response", "content": "your\_response\_content"\}\}\\
The "content" needs to be as concise as possible.\\
\bottomrule
\end{tabular}
\caption{Prompt template for Member third input on the Multi-document Question Answering Task.}
\label{prompt:multi-doc_member_third}
\end{table}
\noindent\textbf{Member third output}\\

\textit{\{"type": "response", "content": "658,798"\}} \\

\noindent At this point, there is only one cluster left, thus all members have only one unique result.\\
\noindent\textbf{Leader last input}\\
The prompt for the Leader last input is shown in Tabel \ref{prompt:multi-doc_leader_last}. 
At this moment, Leader still retains the dialogue history of the previous two rounds.
\begin{table*}[!ht]
\centering
\begin{tabular}{p{\linewidth}}
\toprule
Here are the responses from all the members. Each member sees different segments of a document, and these segments do not intersect with each other. The correct answer may appear in any one or several members' responses.\\
Note that if a minority of members find information relevant to the question while the majority reply that the document does not contain information relevant to the question, you should pay attention to the replies from those members who found relevant information.\\
\textbf{\# Member Response:} \\
Member 1: 658,798 \\
\\
\textbf{\# Task Description:} \\
Answer the question based on the given passages. Only give me the answer and do not output any other words. \\
\\
\textbf{\# Task Objective:} \\
Did Leshan or Quanzhou  have a population of 658,798 in 2002? \\
\\
\textbf{\# Determination:} \\
Based on the above information, you need to determine if you can solve the task objective. You have two choices:\\
1. If members' responses cannot solve the task objective, or if their responses contain conflicting answers, provide a new instruction for them to answer again.\\
2. Else, if the task objective can be solved, give your final answer as concisely as you can, using a single phrase if possible. Do not provide any explanation.\\
Your output must following the JSON format: \{\{"type": "answer", "content": "your\_answer\_content"\}\} or \{\{"type": "instruction", "content": "your\_instruction\_content"]\}\}\\
\bottomrule
\end{tabular}
\caption{Prompt template for Leader last input on the Multi-document Question Answering Task.}
\label{prompt:multi-doc_leader_last}
\end{table*}\\
\noindent\textbf{Leader last output}\\
Leader integrates the answers to the two sub-problems to obtain the final answer to the original problem.\\

\textit{\{"type": "answer", "content": "Quanzhou"\}}\\

\subsection{Retrieve}
The Retrieve-type tasks are divided into three types: Retrieve.KV, Retrieve.PassKey, and Retrieve.Number. Although the task descriptions vary for different tasks, the interaction trajectories are similar, and the prompt template is also the same. Therefore, Retrieve.KV task is chosen here as an example for demonstration.\\
\noindent\textbf{Leader first input}\\
First, the leader recruits members according to the task description as shown in Table \ref{prompt:single-doc_leader_first}.\\

\noindent\textbf{Leader first output}\\
\textit{\{"type": "member", "content": "KV member"\}}\\

\noindent\textbf{Leader next input}\\
The prompt for the Leader next input is shown in Tabel \ref{prompt:KV_leader_next}.\\ 
The content of \#Task Description and \#Task Objective is derived from user input.\\
\begin{table}[!ht]
\centering
\begin{tabular}{p{\linewidth}}
\toprule
You are the leader of a team of \{member\_nums\} members. Your team will need to collaborate to solve a task. The rule is:\\
1. Only you know the task description and task objective; the other members do not.\\
2. But they will receive different documents that may contain answers, and you need to send them an instruction to query their document.\\
3. Your instruction need to include your understanding of the task and what you need them to focus on. If necessary, your instructions can explicitly include the task objective.\\
4. Finally, you need to complete the task based on the query results they return.\\
\textbf{\# Task Description:} \\
Finding the corresponding value from a dictionary and a key. \\
\\
\textbf{\# Task Objective:} \\
Key: "2b114db0-d87e-42d2-9b4c-0b1f115976ad"\\
The value associated with the specified key is: \\
\\
\textbf{\# Generate Instruction for Members:} \\
Now, you need to generate an instruction for all team members. You can ask them to answer a certain question, or to extract information related to the task, based on their respective documents.\\
Your output must following the JSON format: \{\{"type": "instruction", "content": "your\_instruction\_content"\}\}\\
\bottomrule
\end{tabular}
\caption{Prompt template for Leader next input on the Retrieve.KV Task.}
\label{prompt:KV_leader_next}
\end{table}\\
\noindent\textbf{Leader next output}\\
\textit{\{
    "type": "instruction", 
    "content": "Finding the corresponding value from a dictionary and a key.
    Key: "2b114db0-d87e-42d2-9b4c-0b1f115976ad"
    The value associated with the specified key is: "
\}}\\
\noindent\textbf{Member first input}\\
The prompt for the Member first input is shown in Tabel \ref{prompt:KV_member_first}.\\
The content of \#Document  is a part of the complete context, with variations for each member.\\
The content of \#Instruction originates from the first output of the leader.\\
\begin{table}[!ht]
\centering
\begin{tabular}{p{\linewidth}}
\toprule
\textbf{\# Document:} \\
\graytexttt{\{member\_document\}}\\
\\
\textbf{\# Instruction:} \\
Finding the corresponding value from a dictionary and a key.\\
Key: "2b114db0-d87e-42d2-9b4c-0b1f115976ad"\\
The value associated with the specified key is: \\
\\
You are an experienced writer; please summarize the content in the document related to the instructions in a <scratchpad> tag, then describe your response."
Your output must following the JSON format: \{\{"type": "response", "content": "your\_response\_content"\}\}\\
The "content" needs to be as concise as possible.\\
\bottomrule
\end{tabular}
\caption{Prompt template for Member first input on the Retrieve.KV Task.}
\label{prompt:KV_member_first}
\end{table}\\
\noindent\textbf{Member first output}\\
Except for the members who refuse to answer, member 1, member 9, and member 12 provide three different responses. \\
\textit{Clusters:\\ \{'f1ec3dd8-b6e7-4af5-a589-fda0b7693f93': [1], 'd3999999-9999-4999-8999-999999999999': [9], 'd9999999-9999-4999-8999-999999999999': [12]\}}\\

\noindent\textbf{Member next input}\\
The prompt template is the same as the prompt for 'Member next input', while the content of \#Document is the merge of the document from two members.\\
First, merge the documents of member 1 and member 9 to obtain the new result.\\

\textit{Member 1 and Member 9 merge: f1ec3dd8-b6e7-4af5-a589-fda0b7693f93}\\

\noindent Since the merged output is identical to the original output of member 1 but differs from the original output of member 9, member 9 has been removed. Next, merge the documents of member 1 and member 12 to obtain the new result.\\

\textit{Member 1 and Member 12 merge: f1ec3dd8-b6e7-4af5-a589-fda0b7693f93}\\

\noindent Remove member 12 because the merged output is identical to the original output of member 1 and differs from the original output of member 12. At this point, all members have only one unique result. The leader will provide the final answer based on this result.\\
\noindent\textbf{Leader last input}\\
leader summarizes the final answer based on previous information. Details are shown in Table \ref{prompt:KV_leader_last}.\\
\begin{table}[!ht]
\centering
\begin{tabular}{p{\linewidth}}
\toprule
Here are the responses from all the members. Each member sees different segments of a document, and these segments do not intersect with each other. The correct answer may appear in any one or several members' responses.\\
Note that if a minority of members find information relevant to the question while the majority reply that the document does not contain information relevant to the question, you should pay attention to the replies from those members who found relevant information.\\
\textbf{\# Member Response:} \\
Member 1: f1ec3dd8-b6e7-4af5-a589-fda0b7693f93 \\
\\
\textbf{\# Task Description:} \\
Finding the corresponding value from a dictionary and a key. \\
\\
\textbf{\# Task Objective:} \\
Key: "2b114db0-d87e-42d2-9b4c-0b1f115976ad"\\
The value associated with the specified key is:  \\
\\
\textbf{\# Determination:} \\
Based on the above information, you need to determine if you can solve the task objective. You have two choices:\\
1. If members' responses cannot solve the task objective, or if their responses contain conflicting answers, provide a new instruction for them to answer again.\\
2. Else, if the task objective can be solved, give your final answer as concisely as you can, using a single phrase if possible. Do not provide any explanation.\\
Your output must following the JSON format: \{\{"type": "answer", "content": "your\_answer\_content"\}\} or \{\{"type": "instruction", "content": "your\_instruction\_content"]\}\}\\
\bottomrule
\end{tabular}
\caption{Prompt template for Leader last input on the Retrieve.KV Task.}
\label{prompt:KV_leader_last}
\end{table}\\
\noindent\textbf{Leader last output}\\
\textit{{"type": "answer", "content": "f1ec3dd8-b6e7-4af5-a589-fda0b7693f93"}
}\\
\subsection{Math.Find}
\subsubsection{Find the largest number}
\noindent\textbf{Leader first input}\\
First, the leader recruits members according to the task description as shown in Table \ref{prompt:single-doc_leader_first}.\\
\noindent\textbf{Leader first output}\\
\textit{\{"type": "member", "content": "Math member"\}}\\
\noindent\textbf{Leader next input}\\
The prompt for the Leader next input is shown in Tabel \ref{prompt:math_leader_next}.\\
The content of \#Task Description and \#Task Objective is derived from user input.\\
\begin{table}[!ht]
\centering
\begin{tabular}{p{\linewidth}}
\toprule
You are the leader of a team of \{member\_nums\} members. Your team will need to collaborate to solve a task. The rule is:\\
1. Only you know the task description and task objective; the other members do not.\\
2. But they will receive different documents that may contain answers, and you need to send them an instruction to query their document.\\
3. Your instruction need to include your understanding of the task and what you need them to focus on. If necessary, your instructions can explicitly include the task objective.\\
4. Finally, you need to complete the task based on the query results they return.\\
\textbf{\# Task Description:} \\
Finding special integers in a lengthy list. \\
\\
\textbf{\# Task Objective:} \\
You should answer with only one number, no other words. The largest number of the list is: \\
\\
\textbf{\# Generate Instruction for Members:} \\
Now, you need to generate an instruction for all team members. You can ask them to answer a certain question, or to extract information related to the task, based on their respective documents.\\
Your output must following the JSON format: \{\{"type": "instruction", "content": "your\_instruction\_content"\}\}\\
\bottomrule
\end{tabular}
\caption{Prompt template for Leader next input on the Math.Find Task.}
\label{prompt:math_leader_next}
\end{table}\\
\noindent\textbf{Leader first output}\\
\textit{\{"type": "instruction", "content": "You should answer with only one number, no other words. The largest number of the list is: "\}}\\
\noindent\textbf{Member first input}\\
The prompt for the Member first input is shown in Tabel \ref{prompt:math_member_first}.\\
The content of \#Document  is a part of the complete context, with variations for each member.\\
The content of \#Instruction originates from the first output of the leader.\\
\begin{table}[!ht]
\centering
\begin{tabular}{p{\linewidth}}
\toprule
\textbf{\# Document:} \\
\graytexttt{\{member\_document\}}\\
\\
\textbf{\# Instruction:} \\
You should answer with only one number, no other words. The largest number of the list is: \\
\\
You are an experienced writer; please summarize the content in the document related to the instructions in a <scratchpad> tag, then describe your response."
Your output must following the JSON format: \{\{"type": "response", "content": "your\_response\_content"\}\}\\
The "content" needs to be as concise as possible.\\
\bottomrule
\end{tabular}
\caption{Prompt template for Member first input on the Math.Find Task.}
\label{prompt:math_member_first}
\end{table}\\
\noindent\textbf{Member first output}\\
Each member returns the maximum value of the numbers in their documents. The results of each member are recorded and passed to the leader.\\
\noindent\textbf{Leader last input}\\
The prompt for the Leader last input is shown in Tabel \ref{prompt:math_leader_last}.\\
Leader finds the maximum value for the entire document based on the outputs of all members.\\
\begin{table*}[!ht]
\centering
\begin{tabular}{p{\linewidth}}
\toprule
Here are the responses from all the members. Each member sees different segments of a document, and these segments do not intersect with each other. The correct answer may appear in any one or several members' responses.\\
Note that if a minority of members find information relevant to the question while the majority reply that the document does not contain information relevant to the question, you should pay attention to the replies from those members who found relevant information.\\
\textbf{\# Member Response:} \\
Member 1: 94
Member 2: 94
Member 3: 94
Member 4: 94
Member 5: 94
Member 6: 94
Member 7: 94
Member 8: 94
Member 9: 94
Member 10: 94
Member 11: 94
Member 12: 94
Member 13: 94
Member 14: 94
Member 15: 94
Member 16: 94
Member 17: 94
Member 18: 94
Member 19: 94
Member 20: 94
Member 21: 94
Member 22: 94
Member 23: 94
Member 24: 94
Member 25: 93
Member 26: 94
Member 27: 94
Member 28: 94
Member 29: 94
Member 30: 94 \\
\\
\textbf{\# Task Description:} \\
Finding special integers in a lengthy list. \\
\\
\textbf{\# Task Objective:} \\
You should answer with only one number, no other words. The largest number of the list is: \\
\\
\textbf{\# Determination:} \\
Based on the above information, you need to determine if you can solve the task objective. You have two choices:\\
1. If members' responses cannot solve the task objective, or if their responses contain conflicting answers, provide a new instruction for them to answer again.\\
2. Else, if the task objective can be solved, give your final answer as concisely as you can, using a single phrase if possible. Do not provide any explanation.\\
Your output must following the JSON format: \{\{"type": "answer", "content": "your\_answer\_content"\}\} or \{\{"type": "instruction", "content": "your\_instruction\_content"]\}\}\\
\bottomrule
\end{tabular}
\caption{Prompt template for Leader last input on the Math.Find Task.}
\label{prompt:math_leader_last}
\end{table*}\\
\noindent\textbf{Leader last output}\\
\textit{\{"type": "answer", "content": "94"\}
}

\subsubsection{Find the second largest number}
For other tasks in Math.Find, the prompt template remains the same; it will be omitted from here on.\\
\noindent\textbf{Leader first output}\\
\textit{\{"type": "instruction", "content": "You should answer with only one number, no other words. The largest number and second-largest number of the list are: "\}}\\
\noindent\textbf{Member first output}\\
\textit{Member 1: 81, 80\quad\quad\,
Member 2: 81, 80\\
Member 3: 81, 80\quad\quad\,
Member 4: 79, 78\\
Member 5: 81, 80\quad\quad\,
Member 6: 81, 80\\
Member 7: 81, 80\quad\quad\,
Member 8: 81, 80\\
Member 9: 81, 80\quad\quad\,
Member 10: 81, 80\\
Member 11: 81, 80\quad\quad
Member 12: 80, 79\\
Member 13: 81, 80\quad\quad
Member 14: 81, 80\\
Member 15: 81, 80\quad\quad
Member 16: 79, 78\\
Member 17: 81, 80\quad\quad
Member 18: 81, 80\\
Member 19: 81, 80\quad\quad
Member 20: 81, 80\\
Member 21: 81, 80\quad\quad
Member 22: 81, 80\\
Member 23: 81, 80\quad\quad
Member 24: 81, 80\\
Member 25: 81, 80\quad\quad
Member 26: 81, 80\\
Member 27: 81, 80\quad\quad
Member 28: 81, 80\\
Member 29: 81, 80\quad\quad
Member 30: 81, 80}\\
\noindent\textbf{Member first output}\\
\textit{{"type": "answer", "content": "80"}}

\subsubsection{Find the third largest number}
\noindent\textbf{Leader first output}\\
\textit{\{"type": "instruction", "content": "You should answer with only one number, no other words. The largest number, second-largest number and third-largest number of the list are:"\}}\\
\noindent\textbf{Member first output}\\
\textit{Member 1: 86, 85, 84\quad\,
Member 2: 86, 85, 84\\
Member 3: 86, 85, 84\quad\,
Member 4: 87, 86, 85\\
Member 5: 86, 85, 84\quad\,
Member 6: 86, 85, 84\\
Member 7: 86, 85, 84\quad\,
Member 8: 86, 85, 84\\
Member 9: 87, 86, 85\quad\,
Member 10: 87, 86, 85\\
Member 11: 86, 85, 84\quad
Member 12: 86, 85, 84\\
Member 13: 86, 85, 84\quad
Member 14: 88, 87, 86\\
Member 15: 86, 85, 84\quad
Member 16: 87, 86, 85\\
Member 17: 86, 85, 84\quad
Member 18: 86, 85, 84\\
Member 19: 86, 85, 84\quad
Member 20: 86, 85, 84\\
Member 21: 86, 85, 84\quad
Member 22: 86, 85, 84\\
Member 23: 86, 85, 84\quad
Member 24: 86, 85, 84\\
Member 25: 86, 85, 84\quad
Member 26: 88, 87, 86\\
Member 27: 86, 85, 84\quad
Member 28: 86, 85, 84\\
Member 29: 86, 85, 84\quad
Member 30: 86, 85, 84}\\
\noindent\textbf{Member first output}\\
\textit{\{"type": "answer", "content": "80"\}}\\
The method for finding the minimum value is similar to that for finding the maximum value, and it will not be further elaborated here.


\section{Additional Results}
This section complements the experimental results mentioned in the main body of this paper.\\
    \begin{figure*}[t]
    \centering
        \includegraphics[width=\linewidth]{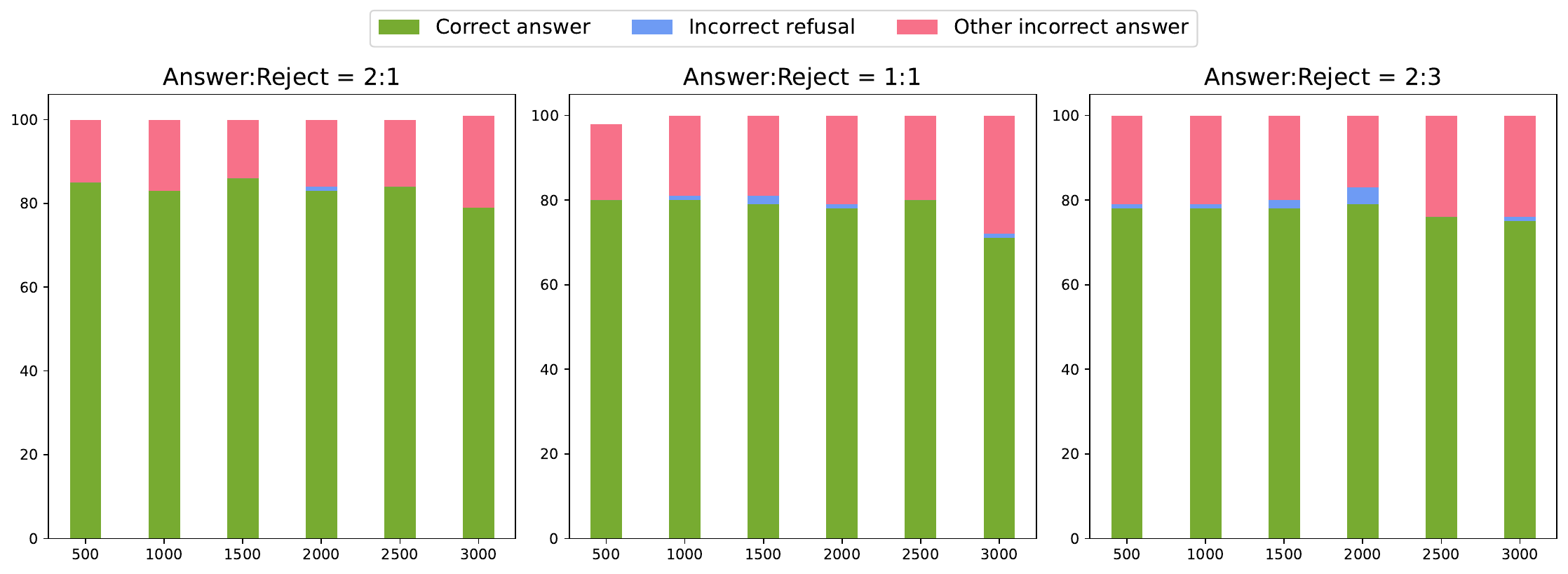}
        \caption{Effect `Reject' type data on model QA capability. `Answer' and `Reject' represent two types of data. For the former, the documents contain answers to questions; whereas for the latter, they do not.}
        \label{fig:hallucination2}
    \end{figure*}

\end{document}